\pdfoutput=1
\documentclass[11pt]{article}

\usepackage{emnlp2021}
\usepackage{stfloats}
\usepackage{url}
\usepackage{verbatim}
\usepackage{enumerate}
\usepackage{bm}
\usepackage{graphicx}
\usepackage{subfigure}
\usepackage{caption}
\usepackage{amsmath}
\usepackage{multirow}
\usepackage{times}
\usepackage{latexsym}

\usepackage[T1]{fontenc}
\usepackage[utf8]{inputenc}

\usepackage{microtype}

\title{Extract then Distill: Efficient and Effective \\Task-Agnostic BERT Distillation}

\author{Cheng Chen\textsuperscript{1}, Yichun Yin\textsuperscript{2}, Lifeng Shang\textsuperscript{2}, Zhi Wang\textsuperscript{3}, Xin Jiang\textsuperscript{2},  Xiao Chen\textsuperscript{2}, Qun Liu\textsuperscript{2}\\ 
 \textsuperscript{1}Department of Computer Science and Technology, Tsinghua University, China \\ 
 \textsuperscript{2}Huawei Noah’s Ark Lab, China \\ 
 \textsuperscript{3}Shenzhen International Graduate School, Tsinghua University, China \\
 \texttt{c-chen19@mails.tsinghua.edu.cn}, \quad \texttt{wangzhi@sz.tsinghua.edu.cn}\\
\texttt{ \{yinyichun,shang.lifeng,jiang.xin,chen.xiao2,qun.liu\}@huawei.com}
}

\begin{document}
\maketitle
\begin{abstract}
Task-agnostic knowledge distillation, a teacher-student framework, has been proved effective for BERT compression. Although achieving promising results on NLP tasks, it requires enormous computational resources. In this paper, we propose \emph{Extract Then Distill} (ETD), a generic and flexible strategy to reuse the teacher's parameters for efficient and effective task-agnostic distillation, which can be applied to students of any size. Specifically, we introduce two variants of ETD, ETD-Rand and ETD-Impt, which extract the teacher's parameters in a random manner and by following an importance metric respectively. In this way, the student has already acquired some knowledge at the beginning of the distillation process, which makes the distillation process converge faster. We demonstrate the effectiveness of ETD on the GLUE benchmark and SQuAD. The experimental results show that: (1) compared with the baseline without an ETD strategy, ETD can save 70\% of computation cost. Moreover, it achieves better results than the baseline when using the same computing resource. (2) ETD is generic and has been proven effective for different distillation methods (e.g., TinyBERT and MiniLM) and students of different sizes. The source code will be publicly available upon publication.
\end{abstract}

\section{Introduction}

With the booming of deep learning in natural language processing (NLP) field, many pre-trained language models (PLMs) are proposed, such as BERT~\cite{bert}, XLNet~\cite{xlnet}, RoBERTa~\cite{roberta}, ALBERT~\cite{albert}, T5~\cite{t5}, ELECTRA~\cite{electra} and so on, achieving state-of-the-art (SOTA) performance on various tasks. However, these large PLMs are computationally expensive and require a large memory footprint, which makes it difficult to execute them on resource-restricted devices.

To tackle this problem, many works propose task-agnostic BERT distillation~\cite{distilbert, tinybert, minilm} to obtain a general small BERT model that can be fine-tuned directly as the teacher model (e.g., BERT-Base) does. However, the process of task-agnostic BERT distillation is also computationally expensive~\cite{structuredqa}. Because the corpus used in the distillation is large-scale, and each training step is computationally consuming that a forward process of teacher model and a forward-backward process of student model need to be performed.

\begin{figure}[t]
\includegraphics[width=0.48\textwidth]{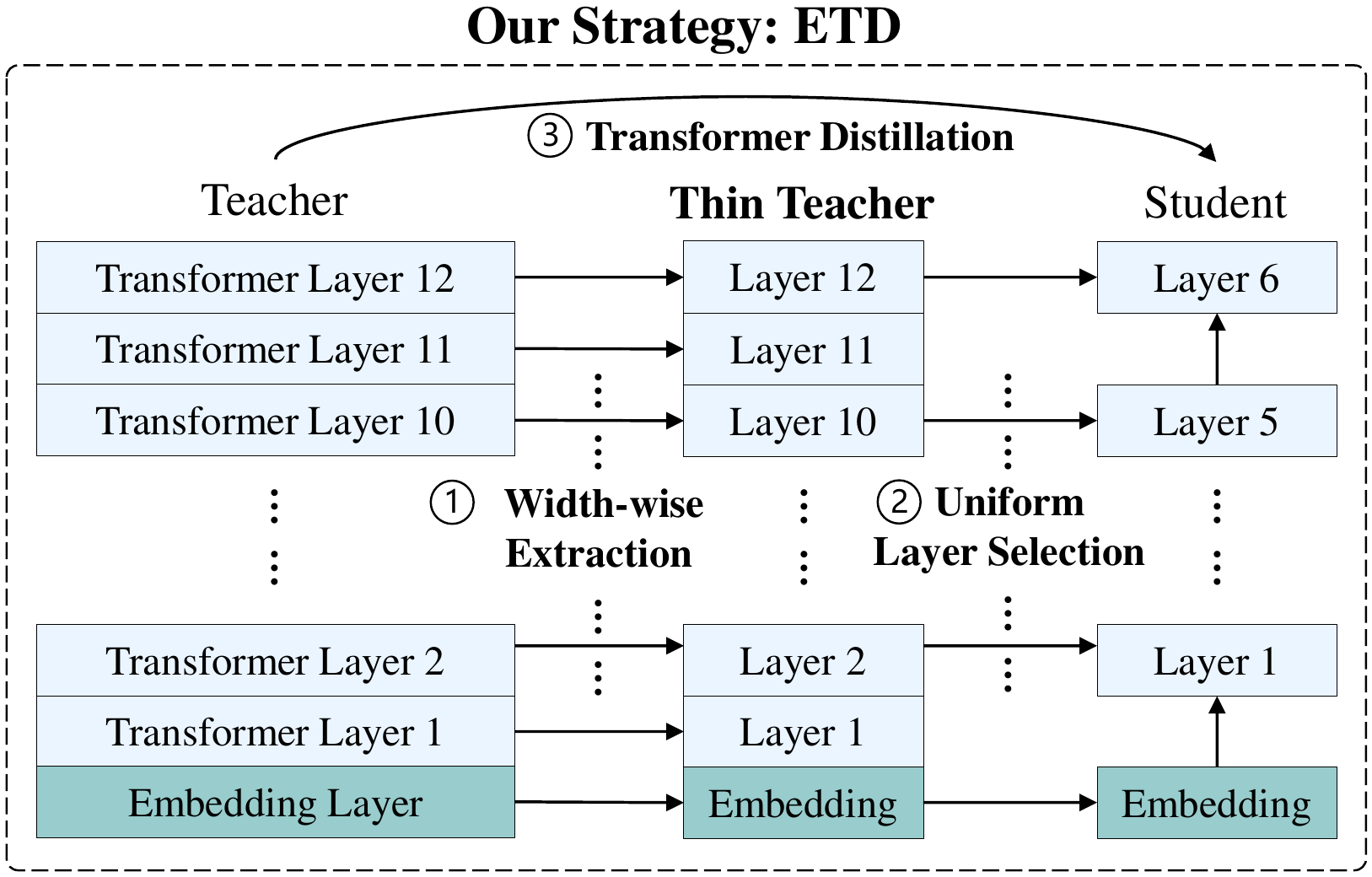}
\caption{Overview of our ETD strategy.}
\centering
\label{fig:etd}
\end{figure}

Reusing the parameters of the teacher model to initialize the student model has been proved effective to improve the efficiency of BERT distillation (e.g., BERT-PKD~\cite{bertpkd} and DistilBERT~\cite{distilbert}). However, these works only do the \emph{depth-wise} extraction that simply copies some Transformer layers [22] from the teacher model, which requires the student to keep the same setting of \emph{width dimensions} including hidden/head/FFN dimension, as the teacher. Actually, in many industrial scenarios, we need to build the student models with different widths and depths to meet various latency and memory requirements.

In this paper, we propose \emph{Extract Then Distill} (ETD) as shown in Fig.~\ref{fig:etd}, a flexible and effective method to reuse the teacher's parameters to initialize the student, which firstly allows the student to have a narrower width than the teacher. Specifically, we propose two methods: ETD-Rand and ETD-Impt, which width-wise extract the teacher's parameters randomly and depending on the importance scores respectively. Then we adopt the strategy of uniform layer selection for depth-wise extraction. Finally, we initialize the student with the extracted parameters, then perform the task-agnostic distillation. 

Our contributions are two-fold: (1) we propose an effective method ETD, which improves the efficiency of task-agnostic BERT distillation by reusing the teacher parameters to the initialize the student model. The proposed ETD method is flexible and applicable to student models of any size. (2) We demonstrate the effectiveness of ETD on the GLUE benchmark and SQuAD. The extensive experimental results show that ETD can save 70\% of computation cost of the baseline, and when using the same computing resources, ETD outperforms the baselines. ETD is general and can be applied to different existing state-of-the-art distillation methods, such as TinyBERT and MiniLM, to further boost their performance. Moreover, the extraction process of ETD is efficient and brings almost no additional calculations.

% The key insight is to view “Reusing teacher model parameters” as a way of knowledge transfer, in this way, the student model already has some priori knowledge at the beginning of the distillation process, therefore, the distillation process will be accelerated.

\section{Preliminary}
\subsection{Architecture of BERT}
\noindent {\bf Embedding Layer.} Through embedding layer, all tokens in a sentence are mapped to vectors of the \textit{hidden size} $\{\bm{x}_i\}^{|x|}_{i=1}$, where $|x|$ means the number of tokens in this sentence. A special token \textsc{[CLS]} is added at the beginning of the sentence for obtaining the representation of the entire sentence.

\noindent {\bf Stacked Transformer Layers.} Transformer layer dynamically encodes context information into the representation vector of each token through the self-attention mechanism, which tackles the problem of semantic ambiguity to a great extent. We denote the output of embedding layer $[\bm{x}_1;...;\bm{x}_{|x|}]$ as \emph{hidden states} $\bm{H}^0$. The stacked Transformer layers compute the contextual vectors as:
\begin{equation}
    H^l = {\rm Transformer}_l(H^{l-1}), l\in \left[1, L\right]
\end{equation}
where L means the number of Transformer layers.

\subsection{Architecture of Transformer}
Each Transformer layer consists of two sub-modules: the multi-head attention (MHA) and the fully connected feed-forward network (FFN). \emph{Residual connection} and \emph{layer normalization} are employed on top of each sub-module.

\noindent {\bf MHA.} In each layer, Transformer uses multiple self-attention heads to aggregate the output vectors of the previous layer. For the $(l+1)$-th Transformer layer, the output of MHA is computed via:
\begin{equation}
\begin{split}
    &\bm{Q}_i=\bm{H}^l\bm{W}_i^Q,\bm{K}_i=\bm{H}^l\bm{W}_i^K,\bm{V}_i=\bm{H}^l\bm{W}_i^V,\\
    &\bm{A}_i=\frac{\bm{Q}_i\bm{K}_i^T}{\sqrt{d_k}},\\
    &{\rm head}_i = {\rm softmax}(\bm{A}_i)\bm{V}_i,\\
    &{\rm MHA}(\bm{H}^l) = {\rm Concat(head_1, ..., head_{\it a})}\bm{W}^O \\
    &\bm{H}^{\rm MHA} = {\rm LayerNorm}(\bm{H}^l + {\rm MHA}(\bm{H}^l)).
\end{split}
\end{equation}

The previous layer’s output $\bm{H}^l$ is linearly projected to queries, keys and values using parameter matrices $\bm{W}^Q, \bm{W}^K, \bm{W}^V$ respectively. ${\rm head}_i$ indicates the context-aware vector which is obtained by the scaled dot-product of queries and keys in the $i$-th attention head. $a$ represents the number of self-attention heads. $d_k$ is the dimension of each attention head acting as the scaling factor, named as head dimension. $d_k \times a$ is equal to the hidden dimension $d_h$. $\bm{H}^{\rm MHA}$ will be the input of the next sub-module FFN.

\noindent {\bf FFN.} It consists of two linear layers and a GeLU~\cite{gelu} function between them:
\begin{equation}
\begin{split}
    &{\rm FFN}(\bm{H}) = {\rm GeLU}(\bm{H}\bm{W}_1 + \bm{b}_1)\bm{W}_2 + \bm{b}_2,\\
    &\bm{H}^{l+1} = {\rm LayerNorm}(\bm{H}^{\rm MHA} + {\rm FFN}(\bm{H}^{\rm MHA})).
\end{split}
\end{equation}

\subsection{Overview of BERT Distillation}\label{subsec:transdil}

There are two widely-used distillation layer mapping strategies, \emph{uniform} and \emph{last-layer}, adopted by TinyBERT~\cite{tinybert} and MiniLM~\cite{minilm} respectively. We adopt the last-layer strategy and follow other settings of TinyBERT, because we empirically find that the last-layer strategy performs better. The last-layer strategy aims to minimize the mean squared errors between the attention patterns $\bm{A}^S$, output hidden states $\bm{H}^S$ of the student’s last Transformer layer and the counterpart $\bm{A}^T,\bm{H}^T$ of the teacher’s last Transformer layer. The overall objective of the last-layer distillation is formulated as:

\begin{equation}
    \mathcal{L}_{\rm KD} = \frac{1}{a} \sum_{i=1}^{a}{\rm MSE}(\bm{A}_i^S, \bm{A}_i^T) + {\rm MSE}(\bm{H}^S\bm{W}, \bm{H}^T),
\end{equation}
where $\bm{A}_i^S$ and $\bm{A}_i^T$ mean the attention matrix corresponding to the $i$-th head in the last layer of student and teacher respectively. The learnable matrix $\bm{W}$ is introduced to solve the problem of mismatched hidden size between student model and teacher model.
\section{Methodology}
Our proposed ETD strategy is general and can be applied to students of any size. It consists of three steps as shown in Fig.~\ref{fig:etd}: (i) \textbf{Width-wise Extraction} that extracts parameters from the teacher to the \emph{thin teacher} by following the principle of \emph{hidden consistency} that will be introduced later. (ii) \textbf{Uniform Layer Selection} that selects layers of the thin teacher with the uniform strategy, and uses these parameters to initialize the student. (iii) \textbf{Transformer Distillation} that performs the last-layer distillation introduced before.

\begin{figure*}[t]
\includegraphics[width=\textwidth]{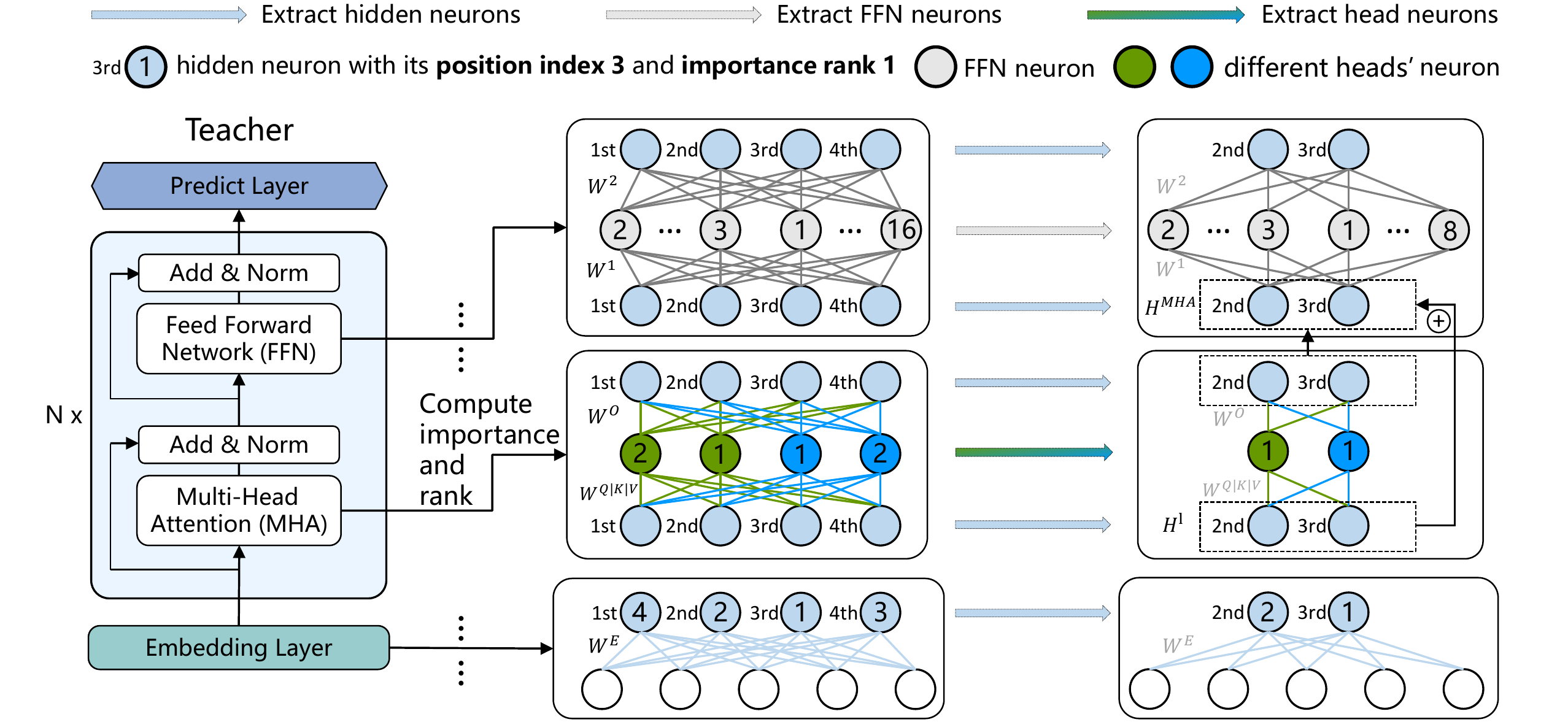}
\caption{Overview of width-wise extraction that includes the extraction of FFN neurons in the intermediate layer of FFN, head neurons in each attention head of MHA and the hidden neurons.}
\centering
\label{fig:pruneprocess}
\end{figure*}

\subsection{Width-wise Extraction.} The width-wise extraction includes the extraction of FFN neurons, head neurons and hidden neurons, which is illustrated in Fig.~\ref{fig:pruneprocess}. Specifically, we refer to the activation units in the intermediate layer of FFN as FFN neurons, which are represented by the gray circles in the figure, and call the activation units in each attention head of MHA as head neurons, which are represented by the green and blue circles in the figure. The hidden neurons, represented by the light blue circles in the figure, refer to the input and output activations of MHA and FFN, and the output of embedding layer. 

Because there are residual connections in the MHA and FFN modules, we argue that the extraction of hidden neurons should follow the principle of {\it hidden consistency} which means that the extracted hidden neurons of different modules/layers should have the same position indexes. For example, as illustrated in Fig.~\ref{fig:pruneprocess}, the hidden neurons with some position indexes of 2 and 3 in different modules are extracted to meet the principle of hidden consistency. If the hidden consistency is not followed, the residual connection relationship will be destroyed, and the knowledge of reused teacher parameters will become chaotic. Experimental results in Table~\ref{tab:main} confirm that if the hidden consistency principle is violated, reusing parameters of the teacher model will even bring negative effects to the student model. In addition, we call the group of hidden neurons whose position index is $i$ in all sub-modules as the $i$-th hidden dimension. Extracting hidden neurons in our ETD strategy can be seen as selecting $d'_h$ hidden dimensions from $d_h$ ones, where $d'_h$ and $d_h$ refer to the hidden size of the student and the teacher respectively. 

Following this principle, we propose two different approaches to extract the teacher's parameters. (i) \textbf{ETD-Rand}: We randomly extract neurons and assign the corresponding weight parameters to student model. (ii) \textbf{ETD-Impt}: We introduce the score-based pruning method in~\cite{sixteen} to extract the relatively important weights from teacher model. Specifically, in ETD-Impt, we calculate the importance of a neuron by the impact on the pre-training loss $\mathcal{L}$ if we remove it. Formally, we denote the $i$-th FFN neuron's value in the intermediate layer of FFN as $\bm{f_i}$. Using the first-order Taylor expansion, it’s importance $I_{f_i}$ can be calculated as:

\begin{equation}
\begin{split}
    I_{f_i} &= \left| \mathcal{L} - \mathcal{L}_{\bm{f_i}=0} \right|, \\
    &= \left| \mathcal{L} - (\mathcal{L}+ \frac{\partial \mathcal{L}}{\partial \bm{f_i}}(0-\bm{f_i}) + R_1(\bm{f_i}))\right|,\\
     &\approx \left|\frac{\partial \mathcal{L}}{\partial \bm{f_i}} \bm{f_i} \right|.
\end{split}
\end{equation}

The item $R_1(\bm{f_i})$ is the remainder of the first-order Taylor expansion so we can ignore it. Similarly, we denote the $i$-th head neuron’s value in a head as $\bm{h_i}$, whose importance $I_{h_i}$ can be calculated by:

\begin{equation}
    I_{h_i} = \left|\frac{\partial \mathcal{L}}{\partial \bm{h_i}}\bm{h_i} \right|.
\end{equation}

As shown in Fig.~\ref{fig:pruneprocess}, we extract the neurons in each head while keeping the number of attention heads. Because attention-based distillation has the constraint that the head number of student should be the same as the teacher. Then, we denote the $i$-th input hidden neuron of MHA module in the $l$-th Transformer layer as $n^l_i$ and the $j$-th hidden dimension as $d_j$. Their importance are denoted as $I_{n^l_i}$ and $I_{d_j}$ respectively, which can be calculated by:
\begin{align}
        \label{algorithn:prune_hidden}
        I_{n^l_i} &= \left|\frac{\partial \mathcal{L}}{\partial \bm{n^l_i}}\bm{n^l_i} \right|,    \\
        \label{algorithn:sum_hidden}
        I_{d_j} &= \sum_{l=0}^{N} I_{n^l_j},
\end{align}
where $\bm{n^l_i}$ means the value of the hidden neuron, $N$ means the number of the teacher's transformer layers. Our preliminary experiments found that summing the neurons’ importance of all Transformer layers in the same hidden dimension is a better choice to obtain $I_{d_i}$. Finally, after calculating the importance scores of all types of neurons, we extract neurons with the highest scores.

According to~\cite{sixteen}, we choose the first 10 thousand sentences in training data
to calculate the importance of neurons.\footnote{The FLOPs (floating-point operations) of calculating the importance score by 10 thousand sentences in training data is 2.5e15, we ignore this computational cost in Table~\ref{tab:main} because of the large difference of magnitude.}

\subsection{Uniform Layer Selection.} After width-wise extraction, we get a thin teacher model, which has the same width with the student. Assuming that the thin teacher and the student has $N$ and $M$ Transformer layers respectively, we choose $M$ layers from the thin teacher model to initialize the student model, and use the set $S$ to represent the chosen layers. It’s necessary to include 0 in $S$, because the index 0 refers to the embedding layer. Formally, the strategy of layer selection we apply is uniform-strategy ($S$ = $\left\{0,\lfloor\frac{N}{M}\rfloor,\lfloor\frac{2N}{M}\rfloor,...,N \right\}$). The other two typical strategies include top-strategy ($S$ = $\left\{0, N\!\!-\!\!M\!\!+\!\!1, N\!\! -\!\!M\!\!+\!\!2, ...,N\right\}$) and bottom strategy ($S$ = $\left\{0, 1, 2, ..., M\right\}$). Our experiments show that the uniform-strategy achieves better results than the other strategies.

\subsection{Transformer Distillation.} We empirically found that the last-layer distillation strategy performs better than the uniform one. Thus, we adopt the last-layer distillation strategy in ETD. As for the performance gap between these two strategies, we can see the experimental results shown in Table~\ref{tab:main} as ``ETD-TinyBERT" (uniform) and ``ETD-Impt" (last-layer).
\section{Experiments}
\subsection{Experimental Setup}

% GD 5轮是一个标准的选项~\ref{tinybert, dynabert}，因为下游任务基本已经收敛。比BERT预训练的轮数要少，主要是因为学生模型较小，不需要过多的训练。

% 我们选出了三个具有代表性的任务，并画出了它们的分数曲线。SST为情感分类任务，MNLI为推断推理任务，SQuAD为阅读理解任务，它们能代表各自的类别的情况。

\noindent {\bf Datasets.} We use English Wikipedia and Toronto Book Corpus~\cite{book} for the distillation data. The maximum sequence length is 512. For evaluation, we use tasks from GLUE benchmark~\cite{wang2019glue} and SQuADv1.1~\cite{2016squad}. We report F1 for SQuADv1.1, Matthews correlation (Mcc) for CoLA and accuracy (Acc) for other tasks.

\noindent {\bf Implementation Details.} We use BERT-Base as our teacher. BERT-Base is a 12-layer Transformers with 768 hidden size. Our ETD method is generic and can be applied to students with varied architectures,to prove it, we instantiate two student models, which are 6-layer and 4-layer model with 384 hidden size respectively. They have different widths with the teacher.

We perform the last-layer distillation based on the released TinyBERT code\footnote{ https://github.com/huawei-noah/Pretrained-Language-Model/tree/master/TinyBERT} and follow the training setting: the batch size and the peak learning rate is set to 128 and 1e-4 respectively, the warm up proportion is set to 10\%. The training epochs of task-agnostic distillation is set to 5 by following~\cite{dynabert, layermapping}.\footnote{The training time that we distill a 6-layer student with 384 hidden size for 5 epochs is (80 hours * 8 V100 cards).} For GLUE fine-tuning, we set batch size to 32, choose the learning rate from \{5e-6, 1e-5, 2e-5, 3e-5\} and epochs from \{4, 5, 10\}. For SQuADv1.1 fine-tuning, we set batch size to 16, the learning rate to 3e-5 and the number of training epochs to 4.

% \footnote{Finally, we fix the epoch number to 4 for all the datasets. Besides, the (learning rate, batch size, maximum sequence length) are different for tasks: (2e-5, 32, 64) for SST-2, (3e-5, 32, 128) for MNLI and (3e-5, 16, 384) for SQuAD v1.1}

% The Main Table
\begin{table*}[t]
\centering
\resizebox{\textwidth}{!}{
\begin{tabular}{l|cc|cccccccc|c}
\hline
Strategy&\#Params&\#FLOPs&SST-2&MNLI&MRPC&CoLA&QNLI&QQP&STS-B&SQuADv1.1&Avg. \\
~ & ~ & (Training) & (Acc) & (Acc) & (Acc) & (Mcc) & (Acc) & (Acc) & (Acc) & (F1/EM) & ~ \\
\hline
BERT-Base(Teacher) &109M &6.43e19 &93.6 &84.7 &87.9 &59.6 &91.6 &91.4 &89.6  &88.4/81.0 &85.85 \\
DistilBERT\ddag\cite{distilbert} &66M & - & 91.3 & 82.2 &87.5 &51.3 &89.2 &88.5 & 86.9 &85.8/77.7 & 82.84 \\
% BERT_{\rm SMALL} & 22M &1.55e19 &88.8 &78.4 &83.4 &42.2 &86.5 &89.3 &87.8  &81.9/73.0 &79.79 \\
\hline
\multicolumn{7}{l}{\it Main Results: Distill the 6-Layer Student Model with 384 Hidden Size}\\
\hline
Rand-Init       &22M &1.55e19 &90.9 &81.6 &86.0 &36.5&89.0&89.3&88.3&85.0/76.6 &80.83   \\
\textbf{ETD-Rand}   &22M &1.55e19 &91.6 &81.7 &\textbf{87.2} &37.7&89.2&89.2&88.1&85.2/76.8 &81.26   \\
\quad \textbf{43\% trained} &22M &6.64e18 &91.0 &81.6 &87.0 &35.9 &89.3 &\textbf{89.4} &88.2 &84.7/76.0 &80.88 \\
\textbf{ETD-Impt}   &22M &1.55e19 &91.3 &\textbf{82.0} &87.1 &\textbf{40.1}&\textbf{89.8}&89.2 & \textbf{88.5}&\textbf{85.6/77.4}& \textbf{81.70} \\
\quad \textbf{28\% trained} &22M &4.43e18 &\textbf{91.9} &81.4 &87.1 &37.2 &89.5 &89.2 &88.4  &84.7/76.1 &81.17 \\
ETD-wo-hidn-con & 22M &1.55e19 &90.6 &81.0 &86.0 &36.1 &89.1 &88.9 &88.2 & 84.9/76.1 & 80.60 \\
\hline
\multicolumn{7}{l}{\it Generality: Combine with Existing BERT Distillation Methods}\\
\hline
Ta-TinyBERT\dag\cite{tinybert}  &22M &1.55e19 &90.6 &80.9 &86.3 &34.9 &87.9 &88.6 &87.2  &83.9/75.2 &80.03 \\
\textbf{ETD-Ta-TinyBERT}  &22M &1.55e19 &91.1 &81.1 &86.0 &38.1 &88.3 &88.6 &87.6  &84.3/75.8&80.64\\
MiniLM\dag\cite{minilm}  &22M &1.55e19 &90.3 &81.4 &84.2 &35.7 &88.8 &89.1 &88.5 &84.5/75.7 &80.30\\
\textbf{ETD-MiniLM}  &22M &1.55e19 &91.1 &82.3 &86.5 &37.5 &89.4 &89.2 &88.4 &84.7/76.1 &81.14\\
\hline
\multicolumn{7}{l}{\it Generality: Distill the 4-Layer Student Model with 384 Hidden Size}\\
\hline 4L-Rand &19M&1.47e19&89.2 &79.2 &85.0 &25.4 & 87.9 &\textbf{88.5} &87.8 & 81.7/72.2 &78.07 \\
\textbf{4L-ETD-Impt} &19M&1.47e19&\textbf{89.6} &\textbf{79.4} &84.5 &\textbf{31.1} & \textbf{88.1} &88.3 &\textbf{88.0} & \textbf{81.9}/\textbf{72.4} &\textbf{78.86} \\
\quad \textbf{35\% trained}  &19M&5.25e18&89.4 &78.9 &\textbf{85.2} &27.4&87.5 &88.2 &87.8 &81.1/71.5 &78.18 \\
\hline
\end{tabular}
}
\caption{Comparison between the performance of student models with different initialization strategies. The fine-tuning results are all averaged over 3 runs on the dev set. \ddag\, denotes that the results are taken from \cite{mobilebert} and \dag\, means the models trained using the released code or the re-implemented code. We report the average score (Avg.) excluding the EM metric of SQuAD.}
\label{tab:main}
\end{table*}

\noindent {\bf Baselines.} As shown in Table~\ref{tab:main}, ``BERT-Base (Teacher)" is the teacher model of our distillation process. Additionally, we also report the results of DistilBERT~\cite{distilbert} as a baseline although it’s not a fair comparison because of the different model parameter amounts (DistilBERT is a 6-layer model whose width remains the same as the teacher model). ``Rand-Init" is our proposed baseline method which initializes the student randomly with the last-layer distillation strategy. The only difference with ETD is that Rand-Init does not reuse the teacher’s parameters. ``ETD-wo-hidn-con" is short for ``ETD-without-hidden-consistency", which denotes it violates the principle of hidden consistency. Specifically, we calculate the importance of the hidden neurons in each layer as shown in Eq.~(\ref{algorithn:prune_hidden}) and extract the most important ones inside each layer directly. Note that it does not guarantee that the position indexes of the extracted hidden neurons in each layer are the same, so it violates the principle. Moreover, in order to verify the generality of our ETD method, we apply ETD-Impt to different typical distillation methods (e.g., TinyBERT and MiniLM). ``TinyBERT" refers to TinyBERT which only does task-agnostic distillation in our paper, and ``ETD-TinyBERT/MiniLM" means that we use the ETD-Impt strategy to initialize the student model. We reproduced MiniLM ourselves because MiniLM doesn't share the model of this architecture setting.\footnote{The results differ from the original paper, possibly because we train with different hyper-parameters.} For a fair comparison, we train them with the same hyper-parameters.

\subsection{Main Results}

\begin{figure*}[t]
\centering
\includegraphics[width=0.9\textwidth]{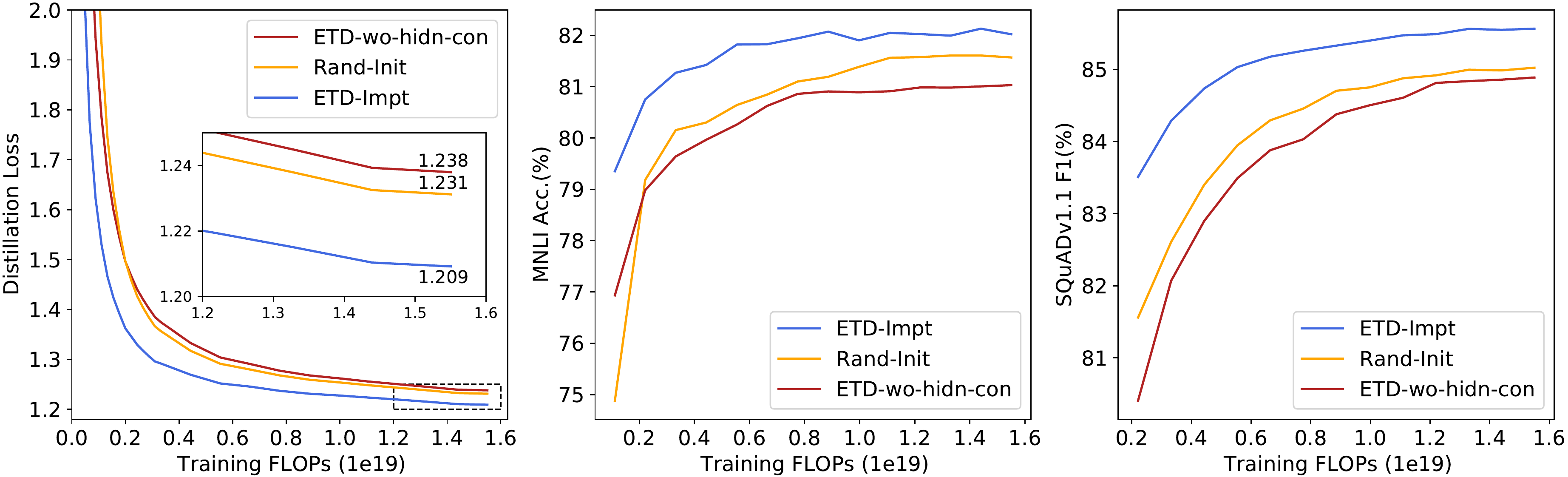}
\caption{The distillation loss curves and the score curves of MNLI and SQuADv1.1.}
\centering
\label{fig:gd_flops_loss}
\end{figure*}

The experimental results in and Table~\ref{tab:main} show that: 1) ETD-Rand/Impt strategies achieve comparable results to Rand-Init, while using only 43\% and 28\% computation cost (FLOPs)\footnote{https://github.com/google-research/electra/blob/master/flops computation.py} respectively. Compared with the baselines of TinyBERT and MiniLM, ETD-Impt can achieve the similar results with even less than 28\% computation cost. 2) ETD-Impt is better than ETD-Rand which indicates that the fine-grained extraction is needed to achieve better results. 3) Reusing the teacher’s parameters is beneficial for most tasks, especially for the tasks of CoLA and MRPC. 4) ETD is generic and can be combined with other distillation methods, such as TinyBERT or MiniLM, to improve their performance. Moreover, ETD works for the 4-layer architecture and achieves comparable results to Rand-Init with only 35\% of the training time. 5) We also find that ETD-Impt can perform comparably to DistilBERT in 7 out 8 tasks and is 3× smaller in size. 6) We demonstrate the effectiveness and necessity of following hidden consistency in Table~\ref{tab:main} and Fig.~\ref{fig:gd_flops_loss}. ``ETD-wo-hidn-con" poses a negative effect on the performance and performs even worse than the ``Rand-Init".

From the distillation loss shown in Fig.~\ref{fig:gd_flops_loss}, we can get the same conclusion that ETD method benefits the distillation efficiency. Moreover, the score curves of the two representative and complex tasks (MNLI and SQuADv1.1) are shown in Fig.~\ref{fig:gd_flops_loss}. We find that lower loss during distillation largely indicates better performance at downstream tasks.
\section{Ablation Study}
In this section, we conduct ablation experiments to analyze the effects of different proposed techniques on the performance, such as with-wise extraction and the strategy of layer selection. Note that all student models are 6-layer with 384 hidden size in this section.

\subsection{The Effect of Each Module}
To understand each type of neuron’s contribution, we also conduct the ablation experiments about the effect of FFN/head/hidden neurons in ETD-Impt. As shown in Table~\ref{tab:ablation}, ``ETD-Impt-rev-ffn" means that we extract the FFN neurons with the lowest importance scores instead of the highest ones. It can test the contribution of FFN neurons and the effectiveness of the importance score based method. It’s the same with ``ETD-Impt-rev-head" and ``ETD-Impt-rev-hidden". Specially, ``ETD-Impt-rev-all" means we reverse the ranks of all types of neurons. The results in Table~\ref{tab:ablation} shown that: 1) Extracting the least important neurons hurts the performance, which means that all three types of neurons have positive effects on the distillation process and the model’s performance in GLUE tasks and SQuAD. It also proves the effectiveness of our extracting method for these three types of neurons. Besides, we can know that among these three types of neurons, FFN neuron’s contribution is the smallest because the gap of distillation loss during convergence between ``ETD-Impt" and ``ETD-Impt-rev-ffn" is the closest. 2) The ETD method is always better than ``Rand-Init", even if it reuses the least important parameters as shown in the comparison between ``ETD-Imptrev-all" and ``Rand-Init", which further proves the effectiveness and necessity of reusing the teacher’s parameters.

\begin{table*}[t]
\centering
\resizebox{0.6\textwidth}{!}{
\begin{tabular}{l|c|cc|c}
\hline
Setting&Distillation Loss&Score&SQuADv1.1&Avg. \\
~ & (Convergence) & ~ & (F1/EM) & ~ \\
\hline
Rand-Init &1.231 & 80.24 & 85.0/76.6 & 80.83 \\
\hline \textbf{ETD-Impt} &\textbf{1.209}&\textbf{81.14}&\textbf{85.6/77.4}&\textbf{81.70}       \\
\hline ETD-Impt-rev-ffn &1.216&80.74&85.1/75.5 &81.29 \\
ETD-Impt-rev-head& 1.218&80.50&85.4/76.7&81.12  \\
ETD-Impt-rev-hidden& 1.219&80.66&85.0/76.6&81.20   \\
ETD-Impt-rev-all&1.227&80.38&84.9/76.3&80.95 \\
\hline
ETD-Impt-top& 1.215&80.55&85.3/76.8&81.14   \\
ETD-Impt-bottom&1.217&80.64&85.2/76.7&81.21 \\
\hline
\end{tabular}}
\caption{Ablation experiments to analyze the effects of different proposed techniques on the performance. The results are validated on the dev set and averaged over 3 runs. Distillation Loss refers to the final loss reached by the model at the end of training. Score refers to the average score of the GLUE tasks. We report the average score (Avg.) excluding the EM metric of SQuAD.}
\label{tab:ablation}
\end{table*}

\subsection{The Strategy of Layer Selection}
We compare the proposed uniform-strategy with the other two strategies: top-strategy (ETD-Impt-top) and bottom-strategy (ETD-Impt-bottom). According to the comparison results as shown in Table~\ref{tab:ablation}, we can find that: 1) The difference between bottom-strategy and top-strategy is not obvious. 2) The uniform strategy is the best performing strategy among them with obvious advantages. The results of the experiment indicate that uniform layer selection is a good strategy because it ensures that the student model can obtain the knowledge from the bottom to the top of the BERT model.
\section{Related Work}
\noindent {\bf Knowledge Distillation for BERT.} Knowledge distillation~\cite{kd} for BERT compression can be categorized into task-agnostic BERT distillation~\cite{distilbert,minilm, mobilebert} and task-specific BERT distillation~\cite{bertpkd, dynabert,tang2019lstm}. Task-agnostic distillation derives a small general BERT with pre-trained BERT as the teacher and large-scale unsupervised corpus as the training dataset. Task-specific BERT distillation learns a small fine-tuned BERT for a specific task with fine-tuned BERT as the teacher and task dataset as the input. TinyBERT~\cite{tinybert} proposes a two-stage distillation framework, including task-agnostic/specific distillation. Our work focuses on the task-agnostic distillation and aims to improve its efficiency. 

% DistilBERT~\cite{distilbert} distills the knowledge from the output distribution of BERT to a student model with 6 layers. MobileBERT~\cite{mobilebert} is a thin version of BERT$_{\rm LARGE}$ while equipped with novel bottleneck structures.  Distilled-BiLSTM$_{\rm SOFT}$~\cite{tang2019distilling} transfers knowledge from BERT to a single-layer BiLSTM~\cite{bilstm}, achieving comparable performance with ELMo~\cite{elmo}.

\noindent {\bf Pruning.} Model extraction is essentially a pruning method. Han~\cite{prune} proposes an iterative pruning method based on weight magnitude. Similarly, the lottery tickets~\cite{ticket} use an iterative method to find the winning lottery ticket. However, these works both require specialized hardware optimizations for acceleration due to the irregular weight sparsity. There are some works of pruning-based Transformer compression~\cite{poorman, sixteen, l0, layerdrop, zhang2020know, dewynter2020optimal, gordon2020compressing}. In this paper, we propose the ETD strategy that adopts pruning techniques to extract FFN/head/hidden neurons of the teacher for finding a better initialization for the student.
 
 %  ~\citet{sixteen} calculates the importance of a head by the impact on loss when we remove it, and prune heads based on their importance scores. Poor man~\cite{poorman} pruned some layers of BERT and fine-tunes it directly, achieving comparable performance with DistilBERT.

\begin{comment}
Recently, Han et al. (2015) propose a
simple compression pipeline, achieving 40 times
reduction in model size without hurting accuracy.
Unfortunately, these techniques induce irregular
weight sparsity, which precludes highly optimized
computation routines. Thus, others explore pruning entire filters (Li et al., 2016; Liu et al., 2017),
with some even targeting device-centric metrics,
such as floating-point operations (Tang et al.,
2018) and latency (Chen et al., 2018). Still other
studies examine quantizing neural networks (Wu
et al., 2018); in the extreme, Courbariaux et al.
(2016) propose binarized networks with both binary weights and binary activations.

However, the high demand for computing resources in training such models hinders their application in practice. In order to alleviate this resource hunger in large-scale model training, we propose a Patient Knowledge Distillation approach to compress an original large model (teacher) into .... 
\end{comment}

\section{Conclusion and Future Work}
In this paper, we proposed a flexible strategy of reusing teacher’s parameters named ETD for the efficient task-agnostic distillation. The experimental results show that our method significantly improves the efficiency of task-agnostic distillation and the fine-grained method ETD-Impt performs better than the random extraction ETD-Rand. In the future, we will optimize the strategy of layer selection and study more advanced pruning methods for ETD, such as L0 penalty~\cite{l0} and group lasso~\cite{grouplasso2006}.

\bibliography{custom}
\bibliographystyle{acl_natbib}

% \appendix

% \section{Example Appendix}
% \label{sec:appendix}

% This is an appendix.

\end{document}